\documentclass{article}
\usepackage[margin=1in]{geometry}
\usepackage{natbib}
\usepackage{authblk}
\usepackage{hyperref}

\date{}
\author[1]{Ferm\'{\i}n Moscoso del Prado Mart\'{\i}n\\
\href{mailto:fm611@cst.cam.ac.uk}{fm611@cst.cam.ac.uk}}
\affil[1]{Department of Computer Science and Technology, University of Cambridge}

\usepackage{amsmath}
\usepackage{amssymb}
\usepackage{booktabs}
\usepackage{times,latexsym}
\usepackage{url}
\usepackage[T2A,T1]{fontenc}
\usepackage[utf8]{inputenc}
\usepackage{graphicx}
\usepackage[spanish,english]{babel}
\usepackage{tikz}
\usepackage{algorithm}
\usepackage{algorithmic}
\usepackage{longtable}
\usepackage{booktabs}
\usepackage{makecell}

\begin{document}

\title{Universal Topological Regularity of Syntactic Structures}

\maketitle

\begin{abstract}
Despite their widespread use, the principles governing the organisation of syntactic dependency trees remain poorly understood. I analyse dependency trees from 124 typologically, genetically, and geographically diverse languages. Their topology departs systematically from randomness. Relative to uniformly sampled random trees, dependency trees exhibit greater structural robustness and lower branching heterogeneity. I propose that these universal regularities emerge naturally from incremental grammatical encoding. I model this process using sublinear preferential attachment. The model accurately reproduces the observed topology. More generally, the results demonstrate how a universal statistical property of syntax can emerge from a simple, cognitively motivated generative process. They further illustrate a broader principle of \emph{efficiency by construction}: communicatively efficient syntactic structures can emerge without direct optimisation for communication.
\end{abstract}

\section{Introduction}
The syntactic structure of human languages is commonly represented using graphs (e.g., \citealp{Chomsky:1957,Tesniere:1959}). In dependency grammar \citep{Tesniere:1959}, words are connected by directed grammatical relations to form a dependency tree representing the syntactic structure of a sentence. Owing to their simplicity, expressive power, and applicability across typologically diverse languages \citep{Tesniere:1959,Nivre:2005}, dependency trees have become a central representation in theoretical linguistics, psycholinguistics, and natural language processing. The availability of large multilingual dependency treebanks \citep{deMarneffe:etal:2021} has made it possible to investigate syntactic organisation quantitatively across hundreds of languages.

Most previous quantitative studies have focused on local properties of dependency trees, including dependency lengths \citep{Gibson:1998,Futrell:etal:2015}, vertex degrees \citep{Ferrer:2004}, dependency direction \citep{Liu:2010}, and edge crossings \citep{Ferrer:2006}. These studies have revealed statistical regularities that are remarkably consistent across languages. In contrast, comparatively little attention has been paid to the graph topology of entire dependency trees \citep{Ferrer:etal:2004}. Whether dependency trees exhibit universal topological regularities beyond these local structural properties remains unknown, although the latter suggest that they might.

To address this question, I develop a quantitative representation of dependency tree topology based on two complementary information-theoretic measures describing structural robustness and branching heterogeneity. This representation provides a common descriptive space for comparing dependency trees across languages and with random trees of the same size. Applying this framework to dependency treebanks from 124 typologically diverse languages reveals that dependency tree topology departs systematically from randomness. Relative to uniformly sampled random trees, dependency trees consistently exhibit greater structural robustness and lower branching heterogeneity, indicating a universal topological organisation that cannot be explained by structural constraints alone.

Having established this empirical regularity, I ask whether it can be explained by a simple generative mechanism grounded in language production. Specifically, I propose that the incremental construction of syntactic structure naturally gives rise to a sublinear preferential attachment process. I show that this simple model accurately reproduces the observed topological organisation across languages and predicts the stretched exponential degree distributions previously reported for dependency trees. More broadly, the results suggest that communicatively efficient syntactic structures can emerge as a consequence of incremental grammatical encoding rather than from direct optimisation for communication.

\section{Measures of Global Dependency Tree Topology}

I characterise global dependency tree topology using two complementary entropy measures. The first quantifies how branching is distributed across vertices, whereas the second captures their overall connectivity of the  trees.

\begin{figure}[t]
\centering
\includegraphics[width=.5\linewidth]{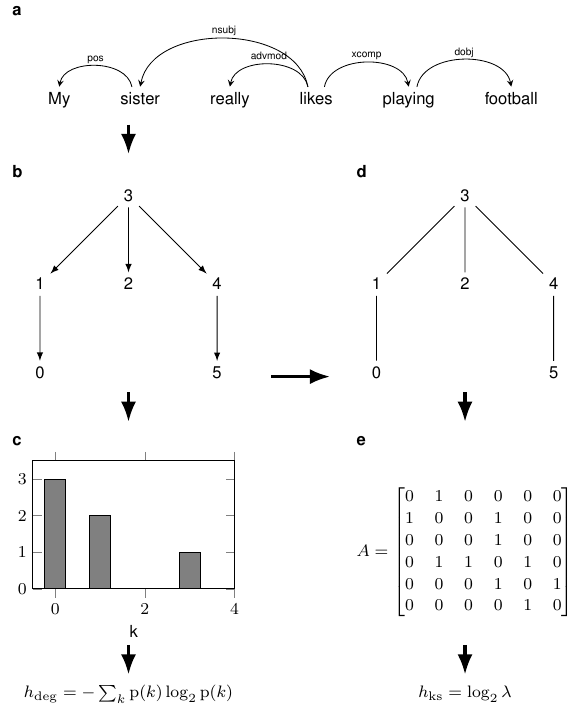}
  \caption{Schema of the computation of the descriptive measures for each dependency graph. {\bf (A)} Dependency graph for an English sentence. {\bf (B)} Digraph skeleton of the dependency graph. {\bf (C)} Distribution of vertex out-degrees used to compute the out-degree entropy ($h_\text{deg}$). {\bf (D)} Undirected skeleton of the digraph. {\bf (E)} Adjacency matrix ($A$) of the undirected skeleton. Its largest positive eigenvalue --its spectral radius ($\lambda$)-- is used for computing the Kolmogorov-Sinai entropy ($h_\mathrm{ks}$) of the graph.}  \label{fig:scheme}
\end{figure}

\subsection{Entropy of a tree's degree distribution}

One important aspect of dependency tree topology is the heterogeneity of branching. Some trees distribute dependents relatively evenly across their vertices, whereas others contain a few highly connected hubs together with many leaves. The heterogeneity of branching in a tree can be quantified by the entropy \citep{Shannon:1948} of the degree distribution of its vertices \citep{Ferrer:Sole:2003a}. Since every non-root vertex has exactly one incoming edge in a dependency tree, topological variability is entirely determined by the distribution of out-degrees. Therefore,  I quantifiy the heterogeneity of the branching structures in the tree using the \emph{entropy of their out-degree distribution} ($h_\mathrm{deg}$). Trees containing both hubs and leaves have high $h_\text{deg}$ values, and trees in which most vertices have similar branching have low $h_\text{deg}$ values.

More formally, for a dependency tree with $N$ vertices, let $k$ be the number of edges that leave a vertex (i.e., the vertex's out-degree). Let $\mathrm{p}(k)$ denote a probability distribution over the values of $k$ in a dependency tree (see Figure~\ref{fig:scheme}C). The entropy of the out-degree distribution is the Shannon entropy over $\mathrm{p}(k)$, 
\begin{equation}
h_\mathrm{deg} = -\sum_{k=0}^{N-1} \mathrm{p}(k) \log_2 \mathrm{p}(k)  \label{eq:hdeg}
\end{equation}

\subsection{Kolmogorov-Sinai Entropy of the trees}

Branching heterogeneity alone does not fully characterise tree topology because trees with similar degree distributions can differ substantially in their global connectivity. I therefore complement $h_\text{deg}$ with the Random Walk Kolmogorov–Sinai entropy ($h_\mathrm{ks}$; \citealp{Demetrius:Manke:2005}), a spectral measure that captures the global connectivity of a tree. This measure is computed from the undirected skeleton of the dependency structure, that is, the pure topology of the dependency tree, not considering the edge directions (see Figure~\ref{fig:scheme}D). This measure has previously been shown to relate to structural robustness of trees (i.e., their capacity to withstand random structural changes)

Formally, for a dependency tree, one can define a $N \times N$ binary adjacency matrix $A$, whose elements $a_{i,j}$ are set to one if vertices $i$ and $j$ are connected (in either direction), and to zero otherwise (see Figure~~\ref{fig:scheme}E). The random walk Kolmogorov-Sinai entropy of the graph is then defined as
\begin{equation}
h_\mathrm{ks} = \log_2 \lambda \label{eq:hks},
\end{equation}
where $\lambda$ denotes $A$'s largest positive eigenvalue (i.e., its spectral radius). 

\subsection{Normalisation of entropy measures}

The values of both entropy measures $h_\mathrm{ks}$ and $h_\mathrm{deg}$ are dependent of the number of vertices ($N$) in a tree. To facilitate comparison of the entropies for sentences of different lengths, I used length-normalised versions of both entropies. For a tree of $N$ vertices (i.e., a sentence of $N$ words) I normalised the measures to their relative values in the $[0,1]$ interval,\footnote{The specific values of the extrema in Equation~\ref{eq:normalisation} are discussed in Appendix~A.}
\begin{equation}
\begin{aligned}
H_\mathrm{ks} & = \frac{h_\mathrm{ks} - \min_N h_\mathrm{ks}}{\max_N h_\mathrm{ks}-\min_N h_\mathrm{ks}}, \\
H_\mathrm{deg} & = \frac{h_\mathrm{deg} - \min_N h_\mathrm{deg}}{\max_N h_\mathrm{deg}-\min_N h_\mathrm{deg}} 
\end{aligned}\label{eq:normalisation}
\end{equation}
The insets in Figure~\ref{fig:resmean} illustrate the interpretation of these normalised measures.

\section{Corpora and Preprocessing}

The analyses were based on the treebanks of the \textit{Universal Dependencies Project v2.11} \citep{deMarneffe:etal:2021}.\footnote{\protect\url{http://hdl.handle.net/11234/1-4923}} For each of the available languages, I concatenated all the treebanks listed for that language. Because both entropy measures are constant for trees with $N \leq 3$ (i.e., all trees with three or fewer vertices are simultaneously line and star graphs), only dependency graphs with at least four vertices were retained.  In addition, to limit processing costs, I discarded any dependency trees with more than 50 vertices (this amounts to excluding fewer than 2\% of the available sentences for any language). Only lexical vertices were retained, with punctuation and range vertices (e.g., ``3--4'' in the CoNLL format) removed. Relation labels were deleted, and vertex labels were replaced by consecutive integers. After these modifications, only those graphs that remained actual trees were retained.

For each language for which there were at least 50 sentences left in the corpus after applying the filters above, I randomly sampled (without replacement) a maximum of 1,000 sentences, if so many were available, or took all available sentences otherwise. This resulted in samples with at least 50 sentences for 124 languages. For each retained dependency graph, I computed the measures $h_\mathrm{deg}$ and $h_\mathrm{ks}$ using Equation~\ref{eq:hdeg} and Equation~\ref{eq:hks}, respectively. Details of the languages considered, their sample sizes and estimated measures are provided in Appendix~B.

    
\subsection{Random tree baselines}

As a baseline, a random directed tree was generated for each dependency tree in the corpus, matching the number of vertices. The baseline trees were sampled uniformly from the space of all directed trees with the same number of labelled vertices. This null model preserves the number of vertices while otherwise removing structural regularities beyond those required for a connected directed tree.

By Cayley's theorem \citep{Cailey:1889}, there are exactly $N^{N-2}$ labelled trees on $N$ vertices. Each labelled tree is in one-to-one correspondence with a \emph{Prüfer sequence} \citep{Prufer:1918}: a unique sequence of length $N-2$ whose elements are the vertex labels. A directed tree is obtained by selecting one of the $N$ vertices as the root, which uniquely determines the orientation of every edge. Consequently, each directed tree is uniquely represented by an \emph{extended Prüfer sequence}, consisting of a standard Prüfer sequence with the identity of the root vertex appended to it. Uniform sampling of directed trees therefore reduces to uniform sampling from the set of extended Prüfer sequences of length $N-1$.

Using this procedure, one random directed tree was generated for each dependency tree, with the number of vertices matched exactly. The quantities $h_\mathrm{deg}$ and $h_\mathrm{ks}$ were then computed for the random trees using Equations~\ref{eq:hdeg} and~\ref{eq:hks}, in the same manner as for the dependency trees.

\section{Universal Topological Organisation of Syntactic Dependency Trees} 

\begin{figure}[t]
\centering
  \includegraphics[width=0.7\linewidth]{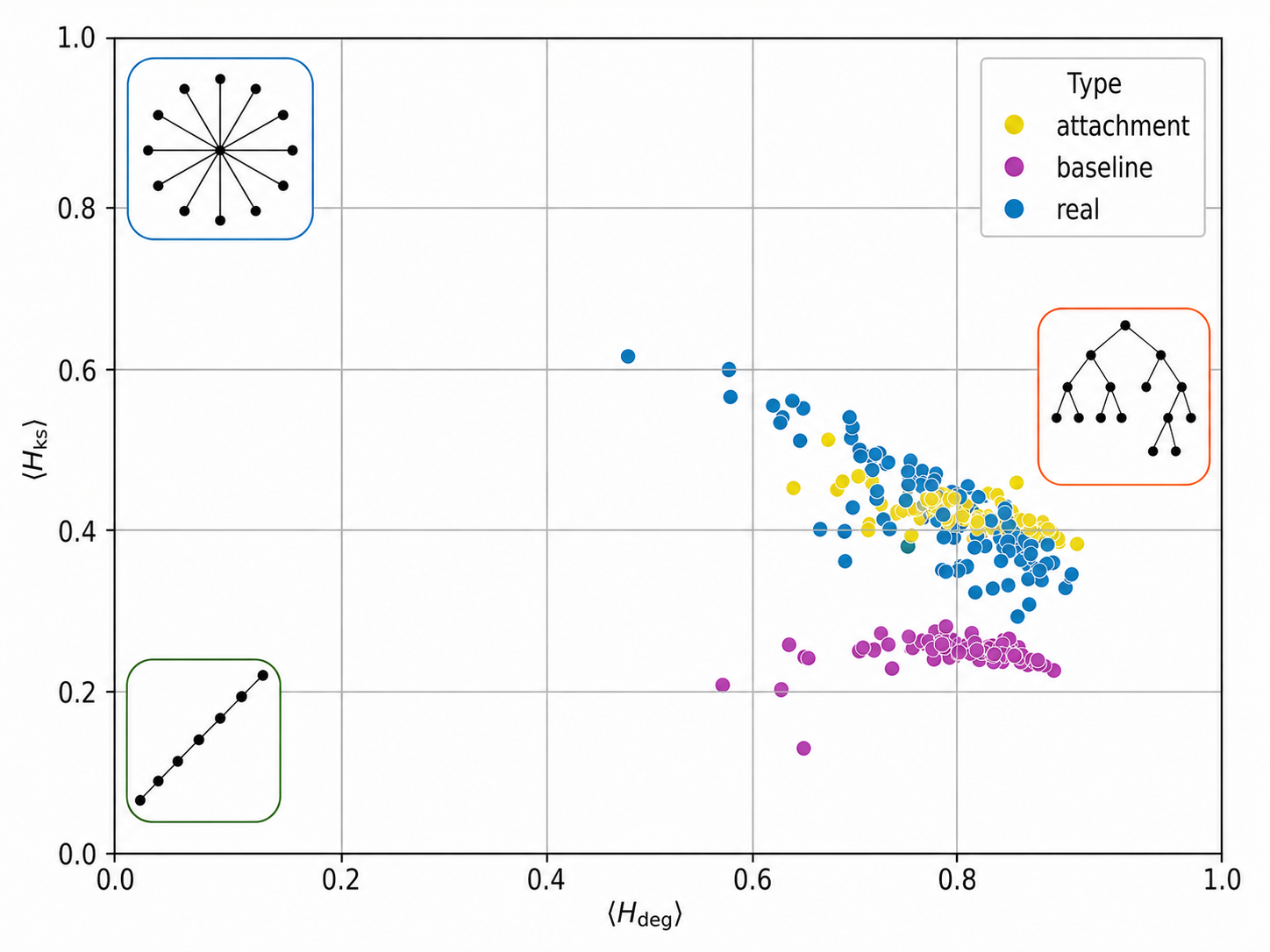}
  \caption {Mean normalised topological measures for each language. Blue: dependency trees from the Universal Dependencies corpora. Purple: uniformly sampled random trees matched for sentence length. Yellow: trees generated by the sublinear preferential attachment model, also matched for sentence length. The inset trees illustrate the interpretation of the two topological measures: path trees have low structural robustness and branching heterogeneity, star trees have high structural robustness and low branching heterogeneity, whereas heterogeneously branching --complex-- trees have high branching heterogeneity.}
  \label{fig:resmean}
\end{figure}

Figure~\ref{fig:resmean} shows the mean normalised measures $H_\text{deg}$ and $H_\text{ks}$ for each of the 124 languages (blue), together with the corresponding values for matched uniformly sampled random trees (purple). Dependency trees differ systematically from the random baseline: Across all 124 languages, $H_\text{ks}$ is significantly higher than expected under the baseline (paired $t[123]=31.25,\,p<.0001$), with all 124 languages showing the same direction of the effect. This is paired with a slight decrease in the average values of $H_\mathrm{deg}$ (paired $t[123]=-3.84,\,p=.0002$, two-tailed), which holds for the majority (60\%) of languages. In short, compared with uniformly sampled trees of the same size, dependency trees  consistently exhibit substantially greater structural robustness, showing also a modest reduction in branching heterogeneity.

\begin{figure}[t]
\centering
\includegraphics[width=0.9\linewidth]{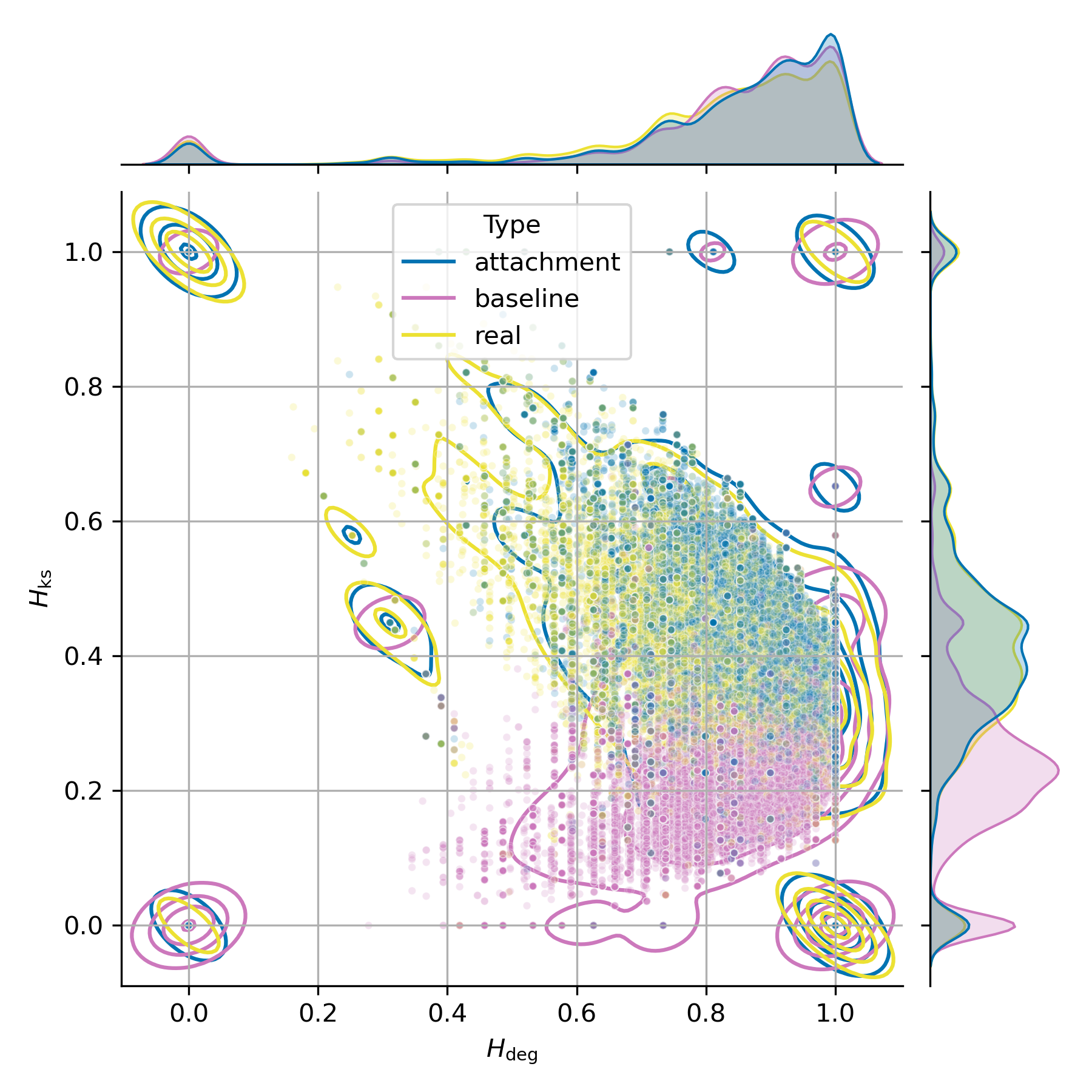}
  \caption {Normalised topological measures for individual dependency trees. Each dot represents one tree: \textit{blue}, real dependency trees; \textit{purple}, uniformly sampled random trees matched for sentence length; and \textit{yellow}, trees generated by sublinear preferential attachment (also matched for sentence length). Contours and marginal distributions show kernel density estimates.}
  \label{fig:resindiv}
\end{figure}

The distinction between real and uniformly sampled trees is sufficiently pronounced that individual dependency trees (see Figure~\ref{fig:resindiv})-—not only language-level averages—- can be classified as real or random with 79\% accuracy using only $h_\text{ks}$ and $h_\text{deg}$ in a simple logistic classifier (see Appendix~C). This classification accuracy goes up to 85\% when one considers sentences with at least ten words, for which a much larger diversity of graph topologies is possible.

\section{Incremental Production Explains the Universal Topological Organisation}

\subsection{Incremental production and sublinear preferential attachment}

Human language production proceeds incrementally, with syntactic structures assembled one word (or chunk) at a time rather than from a fully specified sentence plan \citep{Kempen:Hoenkamp:1987,Levelt:1989,Ferreira:Swets:2002,Ferreira:Slevc:2007}. Throughout this process, each lexical item selected by the \emph{lexical selection process} \citep{Levelt:1989,Bock:Levelt:1994} must be attached by a dependency relation to one of the words already present in the partially constructed dependency tree. The choice of attachment point is determined by the intended meaning of the utterance together with numerous syntactic, semantic, pragmatic, and contextual factors that are difficult to model explicitly. Rather than attempting to represent these latent factors directly, one can adopt a probabilistic perspective and continuously estimate, at each stage of production, the relative likelihood that each existing word will attract the next dependency. Since the partial dependency structure is the only observable information available during tree growth, the current number of dependents attached to a word provides a natural observable proxy for its latent propensity to attract additional dependents. Consequently, words that have already accumulated many dependents are estimated to be more likely to receive further ones, naturally giving rise to a rich-get-richer, or \emph{preferential attachment} (PA), process \citep{Barabasi:Albert:1999}.

If syntactic dependency trees were generated by a PA process, one would expect their degree distributions to follow a power law \citep{Ferrer:Sole:2003}. This is, however, known not to be the case. The degree distributions of dependency trees in individual sentences are instead well described by stretched exponential distributions \citep{Ferrer:etal:2004}. The discrepancy likely arises because the assumption that syntactic structures are assembled strictly one word at a time is only an approximation. Function words, for example, may be generated locally during grammatical encoding rather than emerging directly from lexical selection \citep{Levelt:1989,Bock:Levelt:1994}. Moreover, speakers frequently retrieve and reuse partially specified syntactic structures through \emph{syntactic priming}, effectively introducing larger graph fragments or partially specified dependency structures rather than individual words \citep{Bock:1986,Pickering:Ferreira:2008,Branigan:Pickering:2017}. These mechanisms reduce the number of independent attachment events during sentence construction. Consequently, the probability that a vertex acquires additional dependents is no longer determined solely by its current degree; whether it forms part of a larger syntactic structure retrieved as a whole also influences subsequent attachment decisions. The dependence of attachment probability on degree is therefore expected to be weaker than in the purely word-by-word process.

A mathematical generalisation of PA that captures this weaker dependence on degree is the family of models in which the probability of attachment is proportional to $k^\alpha$, where $k$ denotes the current out-degree of a vertex and parameter $\alpha>0$ controls the strength of preferential attachment. Classical (i.e., linear) PA processes corresponds to $\alpha=1$, whereas $\alpha<1$ and $\alpha>1$ define \emph{sublinear} (sPA) and \emph{superlinear} (SPA) preferential attachment, respectively \citep{Krapivsky:etal:2000}. Notably, sPA processes naturally generate stretched exponential degree distributions, matching those observed in syntactic dependency trees. These considerations motivate the hypothesis that the universal topological organisation of dependency trees emerges naturally from an sPA attachment process.

\subsection{Sublinear preferential attachment models dependency structure topology}

\begin{algorithm}[t]
\caption{Construct a random directed tree with $N$ vertices sampled by non-linear preferential attachment with exponent $\alpha$. Returns the set of directed edges in the tree.}\label{alg:sla}
\begin{algorithmic}
\REQUIRE $N > 0$, $\alpha > 0$
\STATE $Edges \Leftarrow \emptyset$
\STATE $Nodes \Leftarrow \{0,1,\ldots,N-1\}$
\STATE $K \Leftarrow$ Array of $N$ integers, with indices starting at $0$.
\STATE $root \Leftarrow$ sample uniformly an element from $Nodes$
\STATE $Linked \Leftarrow \{root\}$
\STATE $Nodes \Leftarrow Nodes - \{root\}$
\STATE $K[root] \Leftarrow 1$
\WHILE{$Nodes \neq \emptyset$}
\STATE $newnode \Leftarrow $ sample uniformly an element from $Nodes$
\STATE $source \Leftarrow$  sample an element $i$ from $Linked$ with probability proportional to $K[i]^\alpha$
\STATE $Nodes \Leftarrow Nodes - \{newnode\}$
\STATE $Linked \Leftarrow Linked \cup \{newnode\}$
\STATE $K[newnode] \Leftarrow 1$
\STATE $K[source] \Leftarrow K[source]+1$
\STATE $Edges \Leftarrow Edges \cup \{(source,newnode)\}$
\ENDWHILE
\RETURN $Edges$
\end{algorithmic}
\end{algorithm}

In graphs generated by sublinear preferential attachment (sPA), the maximum degree among the vertices ($k_{\max}$) scales asymptotically  with the number of vertices $N$ \citep{Krapivsky:etal:2000}, 
\begin{equation}
k_{\max} \approx (\log N)^{\frac{1}{1-\alpha}},
\label{eq:alpha1}
\end{equation}
where $0<\alpha<1$ is the sublinear exponent. Inverting this asymptotic relationship yields the estimator
\begin{equation}
\hat{\alpha}
=
1-\frac{\log\log N}{\log k_{\max}},
\qquad N>1.
\label{eq:alpha2}
\end{equation}
I estimated $\alpha$ for each dependency tree in the corpus using Equation~\ref{eq:alpha2}. For each language, the language-specific exponent was taken as the mean estimate across all of its dependency trees.
Finally, for every dependency tree, I generated a matched sPA tree with the same number of vertices using the corresponding language-specific exponent $\hat{\alpha}_{\mathrm{language}}$. Tree generation followed Algorithm~\ref{alg:sla}.

The yellow dots in Figure~\ref{fig:resmean} show the mean entropy measures of the sPA-generated trees for each language. Unlike the uniformly sampled trees (purple dots), the sPA-generated trees closely match the distribution of the real dependency trees (blue dots). The agreement is even more apparent at the level of individual, length-matched trees (Figure~\ref{fig:resindiv}), whose joint and marginal distributions are nearly indistinguishable. Consistent with this visual overlap, the logistic classifiers (see Appendix~C) can barely distinguish sPA-generated trees from real dependency trees, achieving only 55\% accuracy (58\% for trees containing at least ten vertices). This contrasts with the substantially higher accuracies  obtained when discriminating real dependency trees from uniformly sampled random trees reported in the previous section (i.e., 79\% and 85\%). Although sPA does not fully account for all the processes involved in dependency tree generation, these results indicate that it captures their global topological organisation remarkably well.

\subsection{Generative validation of the sublinear preferential attachment hypothesis}

\begin{figure}
  \centering
  \includegraphics[width=0.8\linewidth]{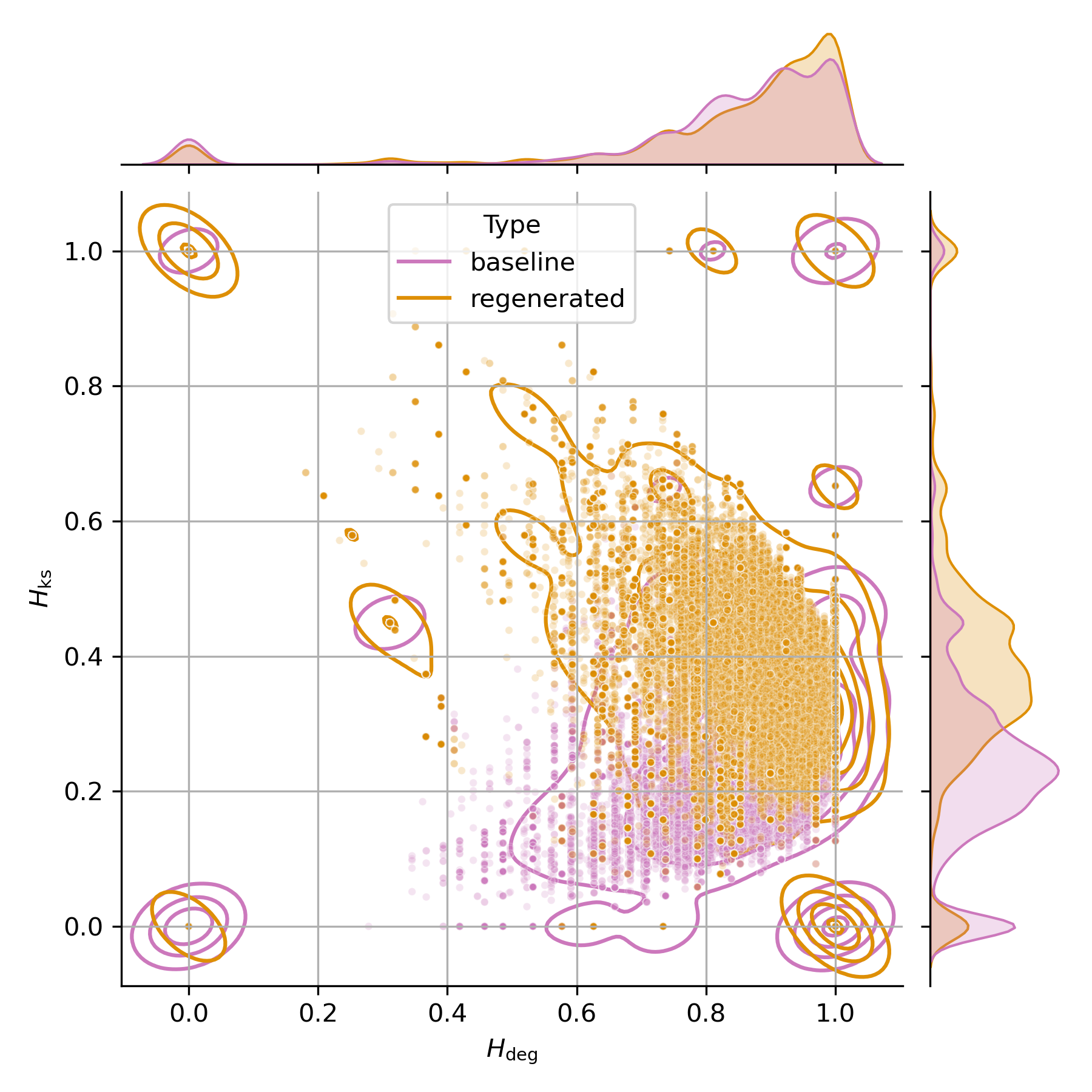}
  \caption{Normalised entropy measures for uniformly sampled random trees (purple) and sPA trees regenerated using the $\hat{\alpha}$ estimates inferred from those random trees (orange). Unlike the regeneration of true dependency trees (Figure~\ref{fig:resindiv}), the regenerated uniform trees exhibit a markedly different distribution. Contours and marginal distributions show kernel density estimates.}
    \label{fig:regenerated}
\end{figure}

Equation~\ref{eq:alpha1} holds asymptotically only for trees actually generated by a true sPA process. Applying the estimator in Equation~\ref{eq:alpha2} to an arbitrary tree will generally produce a value of $\hat{\alpha}<1$, but such an estimate should not be interpreted as evidence that the tree originated from sPA. In other words, Equation~\ref{eq:alpha2} does not constitute a statistical test for sPA. Instead, the validity of the estimated exponent can be assessed through its generative consequences. If a collection of trees has indeed been generated by an sPA process, then new trees generated using the estimated value $\hat{\alpha}$ should reproduce the statistical properties of the original sample. Conversely, if the original trees were not generated by sPA, then trees generated using the estimated exponent will generally exhibit a substantially different distribution.

Regenerating trees by sPA using the exponents estimated from the dependency trees reproduced the original distributions almost exactly (Figure~\ref{fig:resindiv}). In contrast, using the exponents estimated from uniformly sampled random trees produced markedly different distributions (see Figure~\ref{fig:regenerated}). This provides further evidence that the topology of dependency trees is consistent with an underlying sPA process.

\subsection{The precise value of $\alpha$ matters little}

An additional property of the sPA model is illustrated by Figure~\ref{fig:regenerated}. Although the exponents estimated from uniformly sampled random trees fail to regenerate the original random-tree distribution, the resulting sPA trees are nevertheless remarkably similar to those generated using the exponents estimated from real dependency trees (yellow dots in Figure~\ref{fig:resindiv}). This reflects the fact that the sPA model is highly robust to the precise value of $\alpha$. Indeed, any sublinear exponent ($\alpha<1$) yields distributions closely resembling those of real dependency trees.

\begin{figure}
  \centering
  \includegraphics[width=0.7\linewidth]{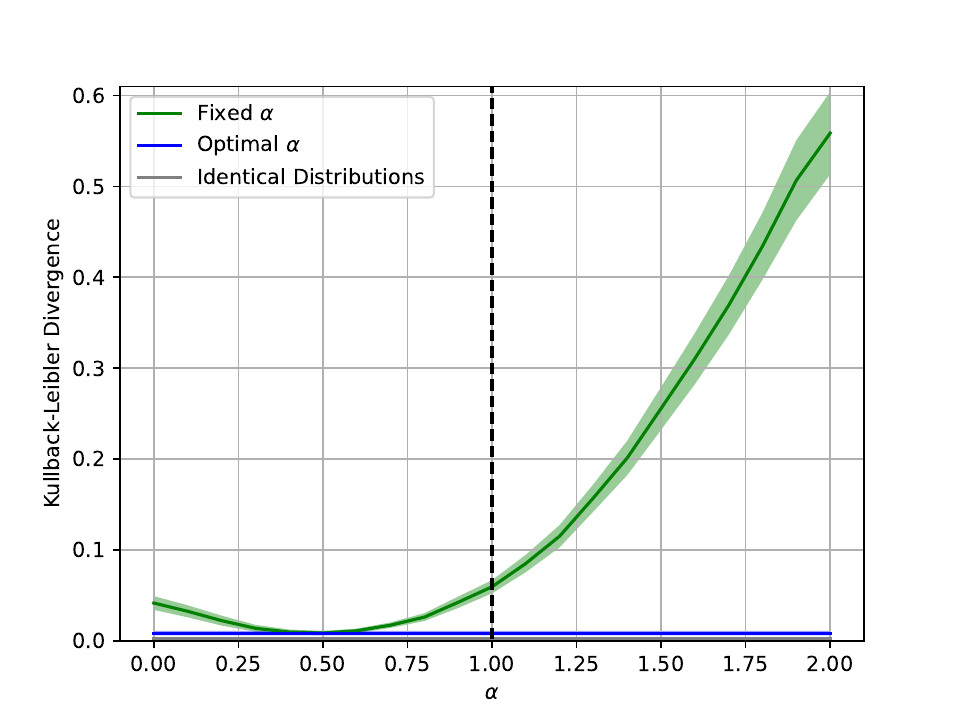}
  \caption{Estimated Kullback--Leibler divergence between the entropy distribution of the real dependency trees for each language and those generated by sublinear preferential attachment using either a fixed value of $\alpha$ for all languages (green) or the language-specific estimated value of $\alpha$ (blue). The grey line denotes the zero-divergence baseline (i.e., estimates for identical distributions), and shaded regions indicate 95\% confidence intervals of the mean. The vertical dashed line separates the sublinear ($\alpha<1$) and superlinear ($\alpha>1$) regimes.}  \label{fig:alphas}
\end{figure}

Figure~\ref{fig:alphas} quantifies this robustness by plotting the estimated Kullback--Leibler divergence (see Appendix~D for estimation method) between the entropy distribution of the real dependency trees and those generated using either the language-specific estimates of $\alpha$ (blue line) or a fixed value of $\alpha$ applied to all trees (green line). Any value of $\alpha$ between $.25$ and $.75$ provides an almost indistinguishable fit, while the remaining sublinear values also yield excellent approximations.\footnote{Neither approach perfectly reproduces the empirical distribution perfectly, as evidenced by the small but consistent gap between both curves and the zero-divergence baseline (grey line), but they are extremely close.} Only in the superlinear regime ($\alpha>1$) do the generated distributions diverge substantially from those observed in human language. These analyses suggest that the defining property of dependency-tree topology is not a particular value of the preferential attachment exponent, but the fact that attachment remains sublinear.


\section{Discussion} 

This study demonstrates that the topology of syntactic dependency trees departs systematically from randomness. Relative to uniformly sampled random trees, dependency trees consistently exhibit substantially greater structural robustness ($h_\mathrm{ks}$) together with slightly lower branching heterogeneity ($h_\mathrm{deg}$). This pattern appears to be universal, holding across a typologically, genealogically, and geographically diverse set of languages spanning multiple registers and modalities. Previous studies have reported topological regularities in \emph{global} syntactic networks, formed by aggregating all dependency relations between word types in a corpus \citep{Ferrer:etal:2004,Liu:2008,Corominas:etal:2009,Corominas:etal:2018}. The present results extend these observations to the level of individual dependency trees.

This universal topological organisation follows naturally from the incremental nature of grammatical encoding, for which ample experimental evidence is available \citep{Levelt:1989,Bock:Levelt:1994}. Word-by-word incrementality is only an approximation, however, since it is occasionally modified by mechanisms such as the separate generation of function words or the reuse of partially specified syntactic structures through syntactic priming \citep{Bock:1986,Pickering:Ferreira:2008,Branigan:Pickering:2017}. Together, these properties of the language production system give rise to a sublinear preferential attachment process \citep{Krapivsky:etal:2000}, which naturally reproduces the observed topological regularities of dependency trees. A direct consequence of sublinearity is that the degree distributions of individual dependency trees should exhibit stretched exponential tails, consistent with previous reports \citep{Ferrer:etal:2004}. This contrasts with the power-law degree distributions observed in global syntactic networks constructed by aggregating dependency relations across an entire corpus \citep{Ferrer:etal:2004,Liu:2008,Corominas:etal:2009,Corominas:etal:2018}.

These findings may also be interpreted from the perspective of communicative efficiency. Relative to the random baseline, dependency trees are more star-like, and star-like graphs have been shown to minimise communicative costs \citep{Ferrer:Sole:2003a}. The present results therefore suggest that these favourable communicative properties may emerge naturally from the production mechanism itself rather than from direct optimisation for communication. More generally, human language is widely believed to avoid gross communicative inefficiency (e.g., \citealp{Moscoso:2013,Pimentel:etal:2021,Caplan:etal:2020}), presumably because evolutionary processes would disfavour markedly inefficient systems \citep{Givon:1991,Croft:Cruse:2004,Hawkins:2004}. At least with respect to syntactic topology, communicative efficiency therefore need not arise through direct optimisation. More generally, the present findings illustrate a broader principle: efficient structures can emerge naturally from simple growth processes, as first emphasised by \citet{Thompson:1917}.

More broadly, these findings help reconcile two influential perspectives on linguistic structure. Psycholinguistic studies have consistently found that speakers prioritise minimising their own production costs over reducing processing costs for listeners when constructing syntactic structures \citep{Levinson:2000,Ferreira:2008}. In contrast, several recent optimisation-based accounts have argued that syntactic structure reflects pressures to maximise communicative efficiency from the listener's perspective (e.g., \citealp{Hahn:etal:2020,Trott:Bergen:2022}). The present findings suggest that these perspectives need not be incompatible. Rather, they support theories proposing that language production mechanisms are the principal determinants of linguistic structure \citep{MacDonald:1999,MacDonald:2013}, while simultaneously showing how listener-oriented efficiency can emerge as a natural consequence of those same mechanisms. In this sense, \emph{efficiency by construction} reconciles the two perspectives: universal topological regularities and favourable communicative properties arise together from the mechanisms by which speakers incrementally construct syntactic structures.

\bibliography{barebones}
\bibliographystyle{compling}

\clearpage

\appendix

\section{Values of the Extrema in Equation~\ref{eq:normalisation}}

\begin{algorithm}[ht]
\caption{Find the maximum value of $h_\mathrm{deg}$ for a given number of vertices $N$.}\label{alg:hdeg}
\begin{algorithmic}
\REQUIRE $N > 1$
\STATE $t \Leftarrow $ array of $N-1$ ones
\STATE $Seen \Leftarrow \{t\}$
\STATE $Stack \Leftarrow \{t\}$
\STATE $best \Leftarrow 0$
\WHILE{$Stack \neq \emptyset$}
\STATE $t \Leftarrow $ \textbf{pop}($Stack$)
\STATE $L \Leftarrow $ \textbf{length}($t$)
\STATE $hist \Leftarrow$ array containing histogram of values in $t$
\STATE $hist[0] \Leftarrow N -L$
\STATE $\mathrm{p} \Leftarrow hist/\sum hist$
\STATE $v \Leftarrow -\sum_{p[i]\neq 0} \mathrm{p}[i] \log_2 \mathrm{p}[i]$
\IF{$v \geq best$}
\STATE $best \Leftarrow v$
\ENDIF
\IF{$L>1$}
\FOR{$i=0$ to $L-1$}
\FOR{$j=i+1$ to $L-1$}
\STATE $t_2 \Leftarrow t$
\STATE $t_2[i] \Leftarrow t_2[i]+t_2[j]$
\STATE Remove $t_2[j]$ from $t_2$
\STATE \textbf{sort}$(t_2)$ 
\IF{$t2 \notin Seen$}
\STATE $Stack \Leftarrow Stack \cup \{t_2\}$
\STATE $Seen \Leftarrow Seen \cup \{t_2\}$
\ENDIF
\ENDFOR
\ENDFOR
\ENDIF
\ENDWHILE
\RETURN $best$
\end{algorithmic}
\end{algorithm}

The extreme values for $h_\mathrm{ks}$ are quite straightforward. On the one hand, $h_\mathrm{ks}$ always takes its maximum for a star graph. The eigenvalues of its adjancency matrix $A$ are the roots of its characteristic polynomial,
\begin{equation}
\det(A -\lambda \cdot I) = 0.
\end{equation}
For star graphs, this determinant can be decomposed recursively into minors, to find that --for any $N$-- it has its maximum positive root at $\sqrt{N-1}$. Therefore:
\begin{equation}
\max_N h_\mathrm{ks} = \frac{1}{2} \log_2(N-1). \label{eq:maxHks}
\end{equation}
Similarly, $h_\emph{ks}$ takes its minimum value for line graphs. It is difficult to find a general closed form of this polynomial's roots for all values of $N$. However, they are easily computed for a specific $N$, so one just finds the largest positive eigenvalue $\lambda$ and then applies Equation~\ref{eq:hks}.

With respect to $h_\mathrm{deg}$, its minimum value is taken for line graphs and star graphs, both of which have all vertices but one having the same degree, therefore,
\begin{equation}
\min_N h_\mathrm{deg} =\log_2 N - \frac{N-1}{N} \log_2(N-1),\label{eq:minHdeg}
\end{equation}
which converges on zero from above for large $N$. On the other hand, the value of $\max_N h_\mathrm{deg}$ is difficult to compute in closed form, as it is related to the integer sum partitions of $N-1$. It can be computed exactly using the Algorithm~\ref{alg:hdeg}. As this algorithm can be slow for large values of $N$, I precomputed the values of the four extrema for $N$ up to 50 before running the remaining simulations.

\section{Languages Analysed}

{\setlength{\tabcolsep}{6pt} 
\begin{longtable}{>{\small}l >{\small}l >{\small}l >{\small}r >{\small}r >{\small}r >{\small}r}
\caption[Language data]{Languages used in the study. Extinct languages, in the sense of not having any remaining native speakers, are marked by $^\dag$. The mean normalised entropy values $\langle H_\mathrm{ks} \rangle$ and $\langle H_\mathrm{deg} \rangle$ are followed by their standard errors.}
\label{tab:langs}\\
\toprule
Language &               Family &               Group & N. Sents &  \thead[r]{Mean Sent.\\Length} & $\langle H_\mathrm{ks} \rangle \pm \mathrm{SE}$ & $\langle H_\mathrm{deg} \rangle \pm \mathrm{SE}$ \\
\midrule
\endfirsthead
\caption[]{\textit{(continued)}} \\
\toprule
Language &               Family &               Group & N. Sents &  \thead[r]{Mean Sent.\\Length} & $\langle H_\mathrm{ks} \rangle \pm \mathrm{SE}$ & $\langle H_\mathrm{deg} \rangle \pm \mathrm{SE}$ \\
\midrule
\endhead
\midrule
\multicolumn{7}{r}{{Continued on next page}} \\
\midrule
\endfoot

\bottomrule
\endlastfoot
Abaza                         &  N.W. Caucasian &                 Abazgi &       86 &                              7.19 &                 $.389 \pm .033$ &                  $.796 \pm .031$ \\
Afrikaans                     &        Indo-European &            Germanic &     1,000 &                             21.64 &                 $.413 \pm .003$ &                  $.813 \pm .004$ \\
Akkadian$^\dag$                      &         Afro-Asiatic &             Semitic &     1,000 &                             12.24 &                 $.305 \pm .006$ &                  $.870 \pm .006$ \\
Akuntsu                       &               Tupian &              Tupari &       93 &                              5.38 &                 $.401 \pm .040$ &                  $.666 \pm .042$ \\
Albanian                      &        Indo-European &            Albanian &       60 &                             13.95 &                 $.358 \pm .013$ &                  $.894 \pm .014$ \\
Amharic                       &         Afro-Asiatic &             Semitic &     1,000 &                              8.66 &                 $.541 \pm .007$ &                  $.631 \pm .008$ \\
\makecell[l]{Ancient\\Greek$^\dag$}                 &        Indo-European &               Greek &     1,000 &                             12.87 &                 $.460 \pm .006$ &                  $.776 \pm .007$ \\
\makecell[l]{Ancient\\Hebrew$^\dag$}                &         Afro-Asiatic &             Semitic &     1,000 &                             21.38 &                 $.387 \pm .002$ &                  $.888 \pm .003$ \\
Apurin\~a                       &             Arawakan &                 Southern &       98 &                              5.90 &                 $.378 \pm .036$ &                  $.751 \pm .037$ \\
Arabic                        &         Afro-Asiatic &             Semitic &     1,000 &                             26.01 &                 $.327 \pm .004$ &                  $.905 \pm .005$ \\
Armenian                      &        Indo-European &            Armenian &     1,000 &                             15.35 &                 $.397 \pm .006$ &                  $.839 \pm .006$ \\
Bambara                       &                Mande &                 Western &      919 &                             11.85 &                 $.441 \pm .006$ &                  $.777 \pm .007$ \\
Basque                        &               \textit{isolate} &                  &     1,000 &                             11.69 &                 $.438 \pm .006$ &                  $.791 \pm .007$ \\
Beja                          &         Afro-Asiatic &            Cushitic &       52 &                             13.06 &                 $.439 \pm .027$ &                  $.791 \pm .034$ \\
Belarusian                    &        Indo-European &              Slavic &     1,000 &                             10.81 &                 $.378 \pm .007$ &                  $.826 \pm .007$ \\
Bhojpuri                      &        Indo-European &               Indic &      343 &                             16.31 &                 $.446 \pm .009$ &                  $.794 \pm .010$ \\
Breton                        &        Indo-European &              Celtic &      807 &                             10.77 &                 $.465 \pm .007$ &                  $.753 \pm .009$ \\
Bulgarian                     &        Indo-European &              Slavic &     1,000 &                             12.45 &                 $.413 \pm .006$ &                  $.817 \pm .007$ \\
Buryat                        &             Mongolic &                 Central &      844 &                              9.60 &                 $.353 \pm .008$ &                  $.803 \pm .009$ \\
Cantonese                     &         Sino-Tibetan &                 Sinitic &      835 &                             12.76 &                 $.554 \pm .008$ &                  $.648 \pm .011$ \\
Catalan                       &        Indo-European &             Romance &     1,000 &                             25.99 &                 $.382 \pm .003$ &                  $.856 \pm .004$ \\
Cebuano                       &         Austronesian &  \makecell[l]{Central\\Philippine} &      150 &                              6.58 &                 $.352 \pm .023$ &                  $.809 \pm .024$ \\
Chinese                       &         Sino-Tibetan &                 Sinitic &     1,000 &                             19.83 &                 $.393 \pm .004$ &                  $.873 \pm .004$ \\
Chukchi                       &  \makecell[l]{Chukotko-\\Kamchatkan} &                 Chukotkan &      653 &                              6.62 &                 $.510 \pm .013$ &                  $.648 \pm .015$ \\
\makecell[l]{Classical\\Chinese$^\dag$}             &         Sino-Tibetan &                 Sinitic &     1,000 &                              5.86 &                 $.346 \pm .010$ &                  $.782 \pm .010$ \\
Coptic$^\dag$                        &         Afro-Asiatic &            Egyptian &     1,000 &                             21.90 &                 $.451 \pm .003$ &                  $.810 \pm .004$ \\
Croatian                      &        Indo-European &              Slavic &     1,000 &                             19.06 &                 $.378 \pm .003$ &                  $.873 \pm .004$ \\
Czech                         &        Indo-European &              Slavic &     1,000 &                             15.52 &                 $.379 \pm .005$ &                  $.844 \pm .006$ \\
Danish                        &        Indo-European &            Germanic &     1,000 &                             16.64 &                 $.442 \pm .005$ &                  $.795 \pm .007$ \\
Dutch                         &        Indo-European &            Germanic &     1,000 &                             14.51 &                 $.436 \pm .005$ &                  $.794 \pm .006$ \\
Emerillon                          &               Tupian &      \makecell[l]{Maweti-\\Guarani} &      132 &                              4.64 &                 $.428 \pm .036$ &                  $.696 \pm .036$ \\
English                       &        Indo-European &            Germanic &     1,000 &                             16.11 &                 $.430 \pm .005$ &                  $.804 \pm .006$ \\
Erzya                         &               Uralic &             Mordvin &     1,000 &                              8.66 &                 $.416 \pm .008$ &                  $.768 \pm .009$ \\
Estonian                      &               Uralic &              Finnic &     1,000 &                             12.40 &                 $.447 \pm .006$ &                  $.767 \pm .008$ \\
Faroese                       &        Indo-European &            Germanic &     1,000 &                             14.64 &                 $.472 \pm .006$ &                  $.764 \pm .007$ \\
Finnish                       &               Uralic &              Finnic &     1,000 &                             10.07 &                 $.435 \pm .007$ &                  $.774 \pm .008$ \\
French                        &        Indo-European &             Romance &     1,000 &                             20.77 &                 $.405 \pm .004$ &                  $.818 \pm .005$ \\
Frisian/Dutch                 &       \textit{code switching} &                  &      385 &                              9.58 &                 $.564 \pm .010$ &                  $.637 \pm .013$ \\
Galician                      &        Indo-European &             Romance &     1,000 &                             29.08 &                 $.356 \pm .003$ &                  $.867 \pm .003$ \\
German                        &        Indo-European &            Germanic &     1,000 &                             16.94 &                 $.423 \pm .004$ &                  $.815 \pm .005$ \\
Gheg                          &        Indo-European &            Albanian &      955 &                             15.65 &                 $.485 \pm .005$ &                  $.753 \pm .006$ \\
Gothic$^\dag$                        &        Indo-European &            Germanic &     1,000 &                             11.03 &                 $.470 \pm .007$ &                  $.751 \pm .008$ \\
Greek                         &        Indo-European &               Greek &     1,000 &                             21.79 &                 $.375 \pm .003$ &                  $.846 \pm .004$ \\
Guajaj\'ara                     &               Tupian &      \makecell[l]{Maweti-\\Guarani} &     1,000 &                              7.15 &                 $.565 \pm .010$ &                  $.584 \pm .011$ \\
Hebrew                        &         Afro-Asiatic &             Semitic &     1,000 &                             22.83 &                 $.342 \pm .003$ &                  $.912 \pm .003$ \\
Hindi                         &        Indo-European &               Indic &     1,000 &                             19.86 &                 $.428 \pm .003$ &                  $.847 \pm .004$ \\
Hindi/English                 &       \textit{code switching} &                  &     1,000 &                             12.71 &                 $.468 \pm .004$ &                  $.778 \pm .005$ \\
Hittite$^\dag$                       &        Indo-European &           Anatolian &      130 &                              9.95 &                 $.481 \pm .020$ &                  $.733 \pm .024$ \\
Hungarian                     &               Uralic &               Ugric &     1,000 &                             20.04 &                 $.382 \pm .004$ &                  $.881 \pm .004$ \\
Icelandic                     &        Indo-European &            Germanic &     1,000 &                             17.80 &                 $.459 \pm .005$ &                  $.777 \pm .006$ \\
Indonesian                    &         Austronesian &     \makecell[l]{Malayo-\\Sumbawan} &     1,000 &                             18.22 &                 $.355 \pm .004$ &                  $.887 \pm .004$ \\
Irish                         &        Indo-European &              Celtic &     1,000 &                             20.04 &                 $.383 \pm .004$ &                  $.868 \pm .005$ \\
Italian                       &        Indo-European &             Romance &     1,000 &                             18.48 &                 $.411 \pm .004$ &                  $.799 \pm .006$ \\
Japanese                      &             Japonic &                 Japanese &     1,000 &                             17.61 &                 $.423 \pm .006$ &                  $.821 \pm .007$ \\
Javanese                      &         Austronesian &            Javanese &      123 &                             14.31 &                 $.364 \pm .012$ &                  $.869 \pm .012$ \\
Kangri                        &        Indo-European &               Indic &      281 &                              7.77 &                 $.513 \pm .014$ &                  $.695 \pm .017$ \\
Karelian                      &               Uralic &              Finnic &      220 &                             11.38 &                 $.422 \pm .012$ &                  $.808 \pm .014$ \\
Karo                          &               Tupian &            Ramarama &      297 &                              4.85 &                 $.615 \pm .023$ &                  $.485 \pm .025$ \\
Kazakh                        &               Turkic &        N.W. &      958 &                              8.49 &                 $.345 \pm .008$ &                  $.800 \pm .008$ \\
K'iche'                         &                Mayan &    \makecell[l]{Greater\\Quichean} &     1,000 &                              6.69 &                 $.321 \pm .009$ &                  $.819 \pm .009$ \\
Komi-Permyak                  &               Uralic &              Permic &       86 &                              8.10 &                 $.438 \pm .029$ &                  $.721 \pm .034$ \\
Komi-Zyrian                   &               Uralic &              Permic &      787 &                             10.24 &                 $.454 \pm .008$ &                  $.763 \pm .009$ \\
Korean                        &               \textit{isolate} &                  &     1,000 &                             11.69 &                 $.330 \pm .006$ &                  $.848 \pm .006$ \\
Kurmanji                      &        Indo-European &             Iranian &      743 &                             12.05 &                 $.429 \pm .005$ &                  $.824 \pm .006$ \\
Latin$^\dag$                         &        Indo-European &              Italic &     1,000 &                             14.10 &                 $.409 \pm .006$ &                  $.828 \pm .007$ \\
Latvian                       &        Indo-European &              Baltic &     1,000 &                             14.35 &                 $.387 \pm .005$ &                  $.843 \pm .006$ \\
Ligurian                      &        Indo-European &             Romance &      275 &                             16.16 &                 $.495 \pm .011$ &                  $.724 \pm .014$ \\
Lithuanian                    &        Indo-European &              Baltic &     1,000 &                             16.11 &                 $.338 \pm .005$ &                  $.878 \pm .005$ \\
Livvi-Karelian                         &               Uralic &              Finnic &      119 &                             10.64 &                 $.401 \pm .017$ &                  $.819 \pm .020$ \\
Low Saxon                     &        Indo-European &            Germanic &       86 &                             23.90 &                 $.440 \pm .012$ &                  $.819 \pm .016$ \\
Maltese                       &         Afro-Asiatic &             Semitic &     1,000 &                             19.31 &                 $.368 \pm .004$ &                  $.861 \pm .005$ \\
Manx                          &        Indo-European &              Celtic &     1,000 &                              7.78 &                 $.289 \pm .006$ &                  $.860 \pm .007$ \\
Marathi                       &        Indo-European &               Indic &      393 &                              7.38 &                 $.412 \pm .015$ &                  $.728 \pm .017$ \\
Mby\'a Guaran\'{\i}                  &               Tupian &      \makecell[l]{Maweti-\\Guarani} &     1,000 &                              9.95 &                 $.447 \pm .006$ &                  $.778 \pm .007$ \\
Moksha                        &               Uralic &             Mordvin &      350 &                              7.77 &                 $.390 \pm .012$ &                  $.787 \pm .014$ \\
Munduruk\'u                     &               Tupian &           Munduruk\'u &      113 &                              6.81 &                 $.401 \pm .031$ &                  $.735 \pm .032$ \\
Naija                         &               \textit{creole} &                  &     1,000 &                             15.55 &                 $.601 \pm .006$ &                  $.584 \pm .008$ \\
Nheengatu                     &               Tupian &      \makecell[l]{Maweti-\\Guarani} &      178 &                              9.62 &                 $.422 \pm .016$ &                  $.771 \pm .021$ \\
North Sami                    &               Uralic &                Sami &     1,000 &                              7.96 &                 $.533 \pm .010$ &                  $.628 \pm .011$ \\
Norwegian                     &        Indo-European &            Germanic &     1,000 &                             15.06 &                 $.451 \pm .006$ &                  $.774 \pm .008$ \\
\makecell[l]{Old Church\\Slavonic$^\dag$}           &        Indo-European &              Slavic &     1,000 &                             10.14 &                 $.493 \pm .008$ &                  $.719 \pm .009$ \\
\makecell[l]{Old E.\\Slavic$^\dag$}               &        Indo-European &              Slavic &     1,000 &                             10.02 &                 $.453 \pm .008$ &                  $.751 \pm .009$ \\
\makecell[l]{Old\\French$^\dag$}                    &        Indo-European &             Romance &     1,000 &                             10.39 &                 $.473 \pm .007$ &                  $.718 \pm .009$ \\
Persian                       &        Indo-European &             Iranian &     1,000 &                             16.61 &                 $.386 \pm .004$ &                  $.867 \pm .004$ \\
Polish                        &        Indo-European &              Slavic &     1,000 &                             10.80 &                 $.391 \pm .007$ &                  $.819 \pm .008$ \\
Pomak                         &        Indo-European &              Slavic &     1,000 &                             12.17 &                 $.499 \pm .007$ &                  $.704 \pm .009$ \\
Portuguese                    &        Indo-European &             Romance &     1,000 &                             15.61 &                 $.405 \pm .005$ &                  $.802 \pm .006$ \\
Romanian                      &        Indo-European &             Romance &     1,000 &                             19.14 &                 $.409 \pm .004$ &                  $.843 \pm .005$ \\
Russian                       &        Indo-European &              Slavic &     1,000 &                             14.45 &                 $.360 \pm .005$ &                  $.861 \pm .006$ \\
Sanskrit$^\dag$                      &        Indo-European &               Indic &     1,000 &                              7.75 &                 $.447 \pm .009$ &                  $.723 \pm .011$ \\
\makecell[l]{Scottish\\Gaelic}               &        Indo-European &              Celtic &     1,000 &                             17.28 &                 $.400 \pm .005$ &                  $.854 \pm .006$ \\
Serbian                       &        Indo-European &              Slavic &     1,000 &                             19.15 &                 $.377 \pm .004$ &                  $.872 \pm .004$ \\
Sinhala                       &        Indo-European &               Indic &      100 &                              7.80 &                 $.433 \pm .017$ &                  $.768 \pm .022$ \\
Skolt Sami                    &               Uralic &                Sami &      216 &                              9.83 &                 $.490 \pm .015$ &                  $.705 \pm .018$ \\
Slovak                        &        Indo-European &              Slavic &     1,000 &                              9.34 &                 $.420 \pm .007$ &                  $.778 \pm .008$ \\
Slovenian                     &        Indo-European &              Slavic &     1,000 &                             16.91 &                 $.435 \pm .005$ &                  $.805 \pm .006$ \\
\makecell[l]{S. Levantine\\Arabic}        &         Afro-Asiatic &             Semitic &       84 &                              7.19 &                 $.345 \pm .026$ &                  $.788 \pm .034$ \\
Spanish                       &        Indo-European &             Romance &     1,000 &                             24.28 &                 $.369 \pm .003$ &                  $.872 \pm .004$ \\
Swedish                       &        Indo-European &            Germanic &     1,000 &                             15.59 &                 $.430 \pm .005$ &                  $.808 \pm .006$ \\
\makecell[l]{Swedish Sign\\Language}         &        \textit{sign language} &                  &      169 &                              9.17 &                 $.483 \pm .019$ &                  $.734 \pm .020$ \\
Swiss German                  &        Indo-European &            Germanic &      100 &                             12.66 &                 $.530 \pm .016$ &                  $.696 \pm .023$ \\
Tagalog                       &         Austronesian &  \makecell[l]{Central\\Philippine} &      178 &                              8.12 &                 $.337 \pm .018$ &                  $.869 \pm .016$ \\
Tamil                         &            Dravidian &                 Southern &      877 &                             11.24 &                 $.410 \pm .009$ &                  $.782 \pm .010$ \\
Tatar                         &               Turkic &        N.W. &      145 &                             12.74 &                 $.336 \pm .012$ &                  $.881 \pm .012$ \\
Telugu                        &            Dravidian &                 S.-Central &      665 &                              4.87 &                 $.398 \pm .015$ &                  $.689 \pm .016$ \\
Thai                          &            Tai-Kadai &                 Tai &      995 &                             21.94 &                 $.380 \pm .003$ &                  $.888 \pm .004$ \\
Tupinamb\'a$^\dag$                     &               Tupian &      \makecell[l]{Maweti-\\Guarani} &      429 &                              7.22 &                 $.436 \pm .013$ &                  $.749 \pm .015$ \\
Turkish                       &               Turkic &        S.W. &     1,000 &                              8.94 &                 $.326 \pm .007$ &                  $.834 \pm .008$ \\
Turkish/German                &       \textit{code switching} &                  &     1,000 &                             15.20 &                 $.544 \pm .005$ &                  $.692 \pm .007$ \\
Ukrainian                     &        Indo-European &              Slavic &     1,000 &                             14.92 &                 $.376 \pm .005$ &                  $.845 \pm .006$ \\
Umbrian$^\dag$                       &        Indo-European &              Italic &       80 &                              8.28 &                 $.346 \pm .029$ &                  $.879 \pm .024$ \\
Upper Sorbian                 &        Indo-European &              Slavic &      635 &                             14.16 &                 $.383 \pm .005$ &                  $.849 \pm .007$ \\
Urdu                          &        Indo-European &               Indic &     1,000 &                             23.30 &                 $.426 \pm .003$ &                  $.845 \pm .004$ \\
Uyghur                        &               Turkic &        S.E. &     1,000 &                              9.80 &                 $.375 \pm .007$ &                  $.816 \pm .007$ \\
Vietnamese                    &       Austro-Asiatic &          Viet-Muong &     1,000 &                             12.86 &                 $.405 \pm .005$ &                  $.838 \pm .006$ \\
Welsh                         &        Indo-European &              Celtic &     1,000 &                             17.88 &                 $.371 \pm .005$ &                  $.848 \pm .006$ \\
W. Armenian              &        Indo-European &            Armenian &     1,000 &                             15.13 &                 $.405 \pm .005$ &                  $.849 \pm .006$ \\
\makecell[l]{W. Sierra\\Puebla Nahuatl} &          Uto-Aztecan &                 Nahuan &      793 &                             10.00 &                 $.418 \pm .008$ &                  $.787 \pm .009$ \\
Wolof                         &          Niger-Congo &   N. Atlantic &     1,000 &                             18.25 &                 $.408 \pm .005$ &                  $.834 \pm .005$ \\
Xibe                          &             Tungusic &      Jurchenic &      736 &                             14.28 &                 $.360 \pm .007$ &                  $.841 \pm .008$ \\
Yakut                         &               Turkic &        N.E. &      237 &                              4.71 &                 $.360 \pm .026$ &                  $.690 \pm .027$ \\
Yoruba                        &          Niger-Congo &              Defoid &      308 &                             21.81 &                 $.420 \pm .006$ &                  $.844 \pm .007$ \\
Yupik                         &         Eskimo-Aleut &                 Eskimoan &      284 &                              7.69 &                 $.440 \pm .011$ &                  $.802 \pm .014$ \\
Zaar                          &         Afro-Asiatic &         W. Chadic &      581 &                              9.92 &                 $.555 \pm .010$ &                  $.617 \pm .013$ \\
\end{longtable}
}

\section{Logistic classifiers}

In order to assess the degree to which it is possible to distinguish between individual real and random graphs I trained four (two distinguishing between real and uniform graphs, and two distinguishing between real and sublinear PA; in each case a model used all available graphs, and the other only graphs with at least ten vertices) binary logistic classifiers on a ramdomly chosen 90\% of the data, and tested it on predicting the remaining 10\%. In each case, one model was trained on all sentences, and the other was trained on sentences having at least ten words. The features used for prediction were the normalised entropy measures $H_\mathrm{ks}$ and $H_\mathrm{deg}$.

\section{Estimation of Kullback-Leibler Divergences}

In order to assess the fit of the joint $h_\mathrm{ks}$ and $h_\mathrm{deg}$ distributions to the real data I use the Kullback-Leibler Divergence (KLD; \citealp{Kullback:Leibler:1951}), between the distribution of actual dependency trees ($\mathrm{p}_\mathrm{real}$) and each generated distribution ($\mathrm{p}_\mathrm{X}$):
\[
D(\mathrm{p}_\mathrm{real}\|\mathrm{p}_\mathrm{X}) = \int_0^\infty\int_0^\infty \mathrm{p}_\mathrm{real}(h_\mathrm{ks},h_\mathrm{deg}) \log \frac{\mathrm{p}_\mathrm{real}(h_\mathrm{ks},h_\mathrm{deg})}{\mathrm{p}_\mathrm{X}(h_\mathrm{ks},h_\mathrm{deg})} d h_\mathrm{ks} \, d h_\mathrm{deg.}
\]
I used a parametric approximation, approximating each distribution by a bivariate Gaussian distribution. For this, I estimated the vector means in each condition,
\[
\boldsymbol\mu = (\langle h_\mathrm{ks} \rangle,\langle h_\mathrm{deg} \rangle)^T,
\]
and the corresponding $2 \times 2$ covariance matrices $\Sigma$. The KLD between two bivariate Gaussians ($P$, $Q$) with vector means $\boldsymbol\mu,\, \boldsymbol m$ and covariance matrices $\Sigma,\, S$, respectively, is calculated using the closed form expression
\[
D(P\|Q) = \frac{1}{2} \left( \mathrm{tr}(S^{-1} \Sigma) -2 + (\boldsymbol m-\boldsymbol\mu) S^{-1} (\boldsymbol m-\boldsymbol\mu) + \log \frac{\det(S)}{\det(\Sigma)}\right).
\]

Even for truly identical distributions, these KLD estimates depend on the sample sizes used for their estimation, and on how well the distributions are modelled by a bivariate Gaussian. In order to provide a `zero' baseline against which to compare the KLD estimates, I used a bootstrap method \citep{Effron:1979}: For each sample of real graphs I computed its KLD to another an equally-sized resampling with replacement from the original sample.

\end{document}